\documentclass[letterpaper, 10 pt, conference]{ieeeconf}  

\usepackage{algorithm,algpseudocode}

\makeatletter
\newcommand\fs@spaceruled{\def\@fs@cfont{\bfseries}\let\@fs@capt\floatc@ruled
  \def\@fs@pre{\vspace{0.75\baselineskip}\hrule height.8pt depth0pt \kern2pt}%
  \def\@fs@post{\kern2pt\hrule\relax}%
  \def\@fs@mid{\kern2pt\hrule\kern2pt}%
  \let\@fs@iftopcapt\iftrue}
\makeatother

\usepackage[utf8]{inputenc}
\usepackage{etoolbox}
\usepackage{gensymb}
\usepackage{fancyhdr}
\usepackage[table]{xcolor}
\usepackage{booktabs}
\usepackage{array} 
\usepackage{multicol}
\usepackage{commath}
\usepackage{amssymb}
\usepackage{siunitx}
\usepackage{graphicx}
\usepackage{xcolor}
\usepackage[normalem]{ulem}
\usepackage{soul}
\usepackage{cite}
\usepackage{algorithm}
\let\labelindent\relax
\usepackage{enumitem}
\usepackage{algpseudocode}
\usepackage[font=small,labelfont=bf]{caption}
\usepackage{cuted}
\usepackage{setspace}
\usepackage[bottom]{footmisc}
\usepackage[colorlinks=true, allcolors=black]{hyperref}

\IEEEoverridecommandlockouts                              

\title{\LARGE \bf
WiSER-X: Wireless Signals-based Efficient Decentralized Multi-Robot Exploration without Explicit Information Exchange}
\author{\textcolor{black}{Ninad Jadhav, Meghna Behari, Robert J. Wood, and Stephanie Gil}
\thanks{\scriptsize{\textcolor{black}{*Authors are affiliated with the John A. Paulson School of Engineering and Applied Sciences, Harvard University, Cambridge, MA 02138, USA. We gratefully acknowledge funding support through Project CETI and NSF CAREER grant CNS-2114733. We thank Hammad Izhar for designing the servo mounts.}}}%
}

\begin{document}
\setstretch{1.0}
\maketitle
\thispagestyle{empty}
\pagestyle{empty}
\begin{strip}
\centering
\vspace{-0.6in}
\includegraphics[scale=0.7]
{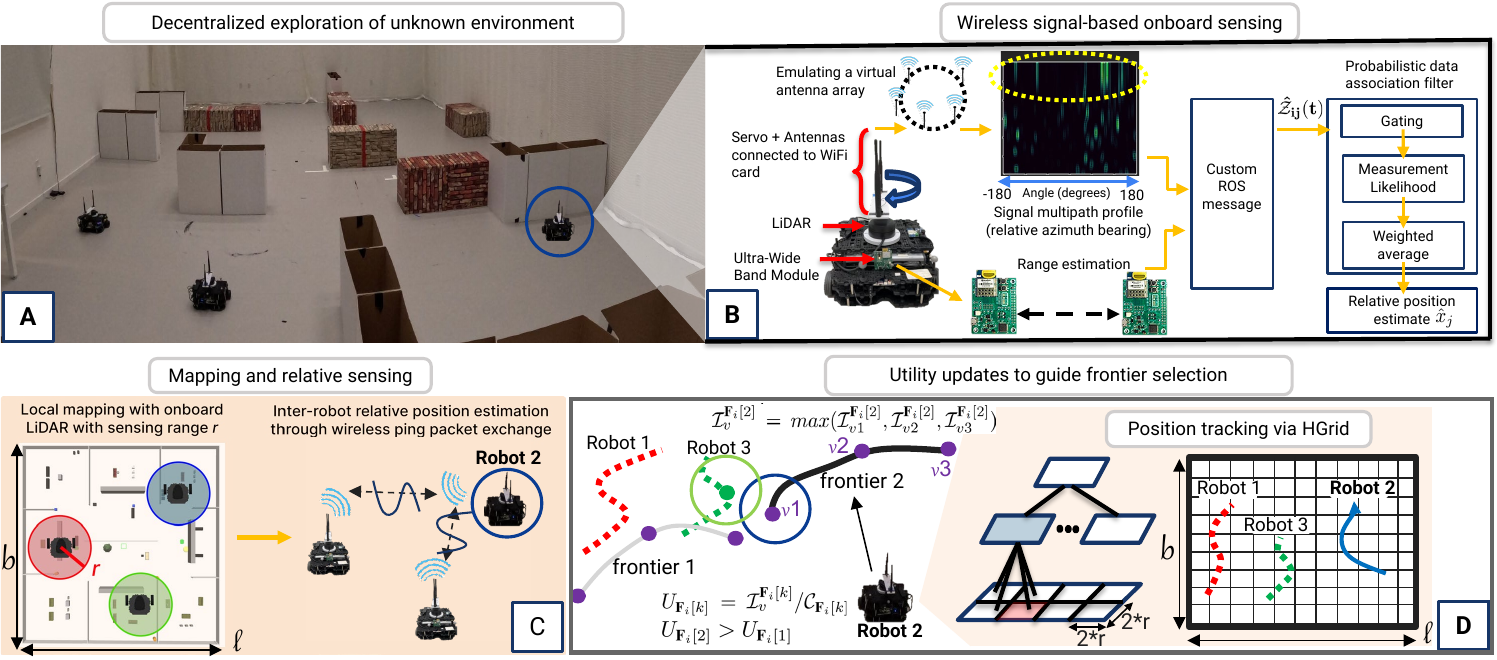} 
  \captionof{figure}{Schematic of the WiSER-X algorithm. (A) Shows an example environment for exploration. (B) Each robot estimates inter-robot relative positions using onboard wireless signal-based sensors; the yellow arc indicates the  signal multipath directions in the angle-of-arrival profile used for bearing estimation. (C) Illustrates how each robot maps its surroundings using LiDAR while exchanging ping packets with neighbors to estimate relative positions. (D) The HGrid structured used for storing the position estimates locally on each robot. The relative position information around a robot's local frontiers enables utility update based on information gain computed a different viewpoints resulting in next frontier selection that minimizes coverage overlaps.}
  \label{fig:intro_fig}
  \vspace{-0.2in}
\end{strip}

\begin{abstract}
We present WiSER-X (Wireless Signal-based Efficient multi-Robot eXploration), a fully decentralized algorithm that enables a team of autonomous robots to coordinate the exploration of unknown environments under severe communication limitations, using only signal ping packets. WiSER-X uses inter-robot relative position estimates, obtained from onboard wireless signal–based sensors, to guide local exploration decisions and minimize redundant coverage, resulting in global coordination from local sensing and decision-making. The algorithm supports asynchronous exploration termination without requiring a shared global map, adapts to heterogeneous robot behaviors, and remains robust to complete robot failures while ensuring full coverage. Simulation results show that WiSER-X achieves 58\% less overlap than a zero-information-sharing baseline and only 23\% more overlap than a full-information-sharing baseline. Hardware experiments further validate the feasibility of WiSER-X using full onboard sensing.\\
\begin{footnotesize}
\noindent Github link: \href{https://anonymous.4open.science/r/wiserx_explore-907D/README.md}{https://anonymous.4open.science/r/wiserx\_explore-907D}
\end{footnotesize}
\end{abstract}
\section{Introduction}
\vspace{-0.05in}
Decentralized coordination among autonomous mobile robots enables faster and more resilient exploration of unknown environments. The efficiency of such coordination often depends on \textit{explicit communication} i.e., inter-robot information exchange, including local maps, shared landmarks, and relative pose estimates. However, this exchange is frequently constrained by limited communication bandwidth, non-line-of-sight (NLOS) conditions, and the computational constraints associated with size, weight, and power limitations of physical robots.

Prior work has made notable progress in addressing the coordination limitations by introducing strategies such as: (1) leveraging prior centralized knowledge to assign robots to specific regions before deployment~\cite{Wurm2008CoordinatedME}, (2) exchanging post-processed data, such as sparse environmental features, to infer shared coverage areas~\cite{vielfaure2021}, and (3) enforcing periodic rendezvous~\cite{Choudhary2017DistributedMW}, such as surfacing for communication or GPS updates during underwater missions~\cite{Kemna2018SurfacingSF}. However, these methods often restrict real-time adaptation to heterogeneous robot performance or failures and can extend mission duration due to the required rendezvous. Ideally, robots could estimate each other’s coverage in real time \emph{without} explicit communication, thereby eliminating high-bandwidth communication requirements for coordination. Such a system would also naturally adapt the exploration workload to individual robot capabilities, improving resilience.

We present WiSER-X, a decentralized coordination algorithm that enhances robots' local frontier-based exploration~\cite{Yamauchi1997AFA} by incorporating relative position estimates of other robots obtained locally (Fig.~\ref{fig:intro_fig}). The \textbf{overarching problem} entails biasing the local exploration strategy of each robot to minimize exploration redundancy. We do so by calculating the information gain of a robots' local frontiers while considering potential coverage overlaps based on the relative positions of the neighboring robots. However, to successfully execute WiSER-X as a decentralized algorithm under communication limitations, we need to address the following key challenges: 

\begin{itemize}[leftmargin=0.125in]
    \item \textbf{Local estimation of relative positions}: Under communication limitations, we derive these relative estimates just from wireless signal pings. The key insight lies in exploiting \emph{implicit information}--- the physical properties of wireless signals such as time-of-flight and phase---to derive range and bearing (Fig.~\ref{fig:intro_fig}B). These principles are broadly applicable across modalities, including acoustic sensing for underwater environments~\cite{Diamant2014LOSAN} and radio-frequency (RF) methods such as WiFi and Ultra-Wideband (UWB)~\cite{Gil2015AdaptiveCI}. Importantly, signal pings needed to obtain these measurements between robots are much more lightweight, typically requiring 64 kB/s (sending approximately 100 ping packets/sec)~\cite{Kumar2014AccurateIL}, can traverse longer distances, and through occlusions. To accurately estimate positions from noisy onboard range and bearing measurements, we use a Probabilistic Data Association Filter (PDAF)~\cite{ShalomPDAF}. WiSER-X also adjusts the weight of a neighboring robot's coverage overlap at a frontier based on the certainty of their relative position estimates.
    \item \textbf{Exploration termination}: During exploration, robots still need to know when to stop without accessing shared map and covering already explored areas. To address this, robots' maintain a history of the relative positions locally in an HGrid, a quadtree like data structure (Fig.~\ref{fig:intro_fig}D), enabling them to track the neighboring robots relative positions and infer overall coverage. Once a robot locally estimates that the environment is sufficiently explored at the global scale, it chooses to visit only those local frontier's where the overlap is below a certain threshold; otherwise they are marked as invalid and no longer a candidate for exploration.
    \item \textbf{Adapting to Heterogeneity and Failures}: WiSER-X also adapts to real-time heterogeneity in robot behavior such as complete failures, minimizing loss of coverage.  
\end{itemize}

We validate our method through extensive simulations and hardware experiments, demonstrating that by relying only on implicit communication, WiSER-X (1) reduces coverage overlap compared to baseline exploration method, (2) faster completion, and (3) effective adaptation to heterogeneous robot behaviors, minimizing missed areas in event of complete failures. Our hardware experiments demonstrate the real-time performance of WiSER-X on mobile ground robots with all onboard sensing.
\section{Related Work}\label{sec:related_work}
\vspace{-0.05in}

Numerous strategies have been proposed to improve exploration of unknown environments using single or multi-robot systems~\cite{Sumer2022,Tao2023}. Decentralized multi-robot exploration typically requires substantial data exchange, including map updates, sensor data, trajectories, or positions within a common reference frame~\cite{fox2006}.
Reducing this \emph{explicit} communication has become a major research focus, with approaches such as map data compression~\cite{psomiadis2023mapcompression,Wu2022MRGMMExplore}, limiting the number of communicating robots~\cite{Unhelkar_Shah_2016}, enabling data sharing only in close proximity~\cite{kulkarni2022,cladera2023}, or designating specific robots to maintain network connectivity~\cite{Jiang2023}. Intermittent rendezvous, where robots periodically meet to exchange information, is another common strategy~\cite{Gao2022,Yu2020CommNetPropagation,Wang2019ActiveRF}. However, all these methods rely on some level of explicit information exchange, can be infeasible in bandwidth-constrained environments and is prone to perceptual aliasing. Despite advances in reducing the communication load, there remains limited research on fully decentralized exploration methods that exploit low-bandwidth \emph{implicit information}. Although some works update the belief state of individual robots to guide exploration~\cite{bramblett2023,Clark2021}, they still depend on communication rendezvous, shared coordinate frame, or centralized “auction”-based coordination~\cite{Smith2018}. To address such constraints, several approaches employ \emph{implicit information} such as incorporating optical cues of relative position into robot coordination strategies. Optical wireless communication has been used for position coordination~\cite{saeed2018underwaterowc,catellaniral2023}, while IR and UV LEDs support robot identification and localization~\cite{Xun2023,Walter2018}. However, these methods depend on line-of-sight.

Another key challenge in decentralized multi-robot exploration of unknown environments is adapting to heterogeneity, including differences in dynamics, sensing capabilities, and potential failures. Methods that pre-assign exploration regions~\cite{Tian2020SearchAR,Fung2019CoordinatingMS} lack flexibility in fully unknown environments due to dependence on prior structural knowledge. Many approaches addressing heterogeneity or failures still rely on centralized control~\cite{Mayya2021, Tziola2023} or periodic explicit communication~\cite{Gao2023, Yan2021}. 

Our method achieves coordinated exploration using only implicit communication while achieving adaptive behavior, by leveraging onboard sensing that employs \emph{wireless signal pings} to locally estimate relative positions of neighboring robots. The use of wireless signal–based sensing in robotics, for example using radio frequency (RF) signals~\cite{WSR_IJRR,Xianjia_2021,Arun2022P2SLAMBB,Wang2024MULANWCML,Cavorsi2022AdaptiveMR,JadhavBhattacharyaSciRob2024} and acoustic signals~\cite{Jabari2016,Jimenez2019ExperimentalRI}, has expanded rapidly in recent years. Recent studies further fuse such sensors to improve relative pose estimation in communication-limited settings~\cite{fishberg2024RAL,shalaby2023,Wang2022WiClosureRA}. We demonstrate that integrating these local position estimates into a decentralized frontier-based exploration framework can enable efficient, low-redundancy exploration, achieving global coordination from local information, without explicit information exchange.

\section{Problem}\label{sec:problem}
\vspace{-0.05in}
We consider a team $\mathcal{R}$ of \emph{n} homogeneous mobile robots exploring an unknown, bounded 2D environment with known dimensions of its outer boundary. Each robot $i\in \mathcal{R}$ is equipped with a finite-range $\ang{360}$ LiDAR that has a scan radius $r$, to map the geometric structure of the environment, represented as a 2D local occupancy grid map $\mathcal{M} \subset \mathbb{R}^2$. A robot uses its local SLAM and frontier-based algorithm for exploration. Given the local map $\mathcal{M}_i$, $\textbf{F}_i$ denotes the set of all frontiers, the boundary of the known and unknown space in $\mathcal{M}_i$~\cite{Yamauchi1998FrontierbasedEU}, generated by robot $i$ at time $t$. $\textbf{F}_i[k]\subset \mathbb{R}^{l\times2}$ denotes the $k^{th}$ frontier consisting of $l$ grid cells.

Robots compute a scalar \emph{utility} value of the $k^{th}$ frontier in $\textbf{F}_i$ based on (1) information gain $\mathcal{I}_{v}^{\textbf{F}_i[k]}$, which is a function of unexplored cells within radius $r$ from a grid cell $v$ (e.g., the center) on a frontier, and (2) the navigation cost $\mathcal{C}_{\textbf{F}_i[k]}$, defined as the path length between the robot's current position and the center grid cell of the frontier. $U_{\textbf{F}_i[k]}$ denotes the utility of $\textbf{F}_i[k]$, representing information gained per unit navigation cost~\cite{Banfi2018StrategiesFC}. $\textbf{U}_{i}$ denotes the set of utilities for all frontiers in $\textbf{F}_i$ at a given timestep.

The autonomous robot team's operation is decentralized, without access to a shared map. Given the communication bandwidth constraints (64 kB/s), the robots cannot rely on real-time inter-robot communication of data such as map updates. Each robot relies solely on the  information collected from its onboard sensors and maintains a local coordinate frame. Robots independently use local path planers for navigation and collision avoidance. A robot can also locally measure the relative range and bearing to other robots, derived from the signal pings. Thus, any robot $i \in \mathcal{R}$ can obtain relative measurements $\hat{x}_{ij}$ to all other robots in its neighborhood $\mathcal{N}_i = \{j | j \in \mathcal{R}, j \neq i \}$. 

We formalize this problem of achieving coordinated exploration, given the limited capabilities of the robots, as follows:

\noindent \textbf{Problem Statement}. \emph{Given the availability of only locally computed inter-robot measurements, develop a frontier-based algorithm such that at every timestep $t \in \{0\hdots\mathcal{T}\}$, where $\mathcal{T}$ denotes total exploration duration, robot $i\in\mathcal{R}$ navigates to a frontier $\mathbf{F}_i[k^{*}]$ in $\mathcal{M}_i$ that has the least overlap with the explored region of other robots $j\in\mathcal{N}_i$}

Solving this problem successfully requires addressing the following three challenges as robots can only rely on implicit communication and do not share any other information.

\noindent\textbf{Challenge 1} \emph{Robot $i$ needs to estimate relative positions $\hat{x}_{ij}$ for all robots $j \in \mathcal{N}_i$ in real-time, in presence of occlusions and beyond visual line-of-sight, by integrating range and bearing measurements obtained from onboard wireless signal-based sensors.}
 
Addressing this challenge enables robot $i$ to evaluate coverage overlaps when robot $j$'s relative positions are in proximity of the robot $i$'s $k^{th}$ frontier $\textbf{F}_i[k]$. Eventually, the robot team needs to stop when the environment is fully explored i.e., when there are no frontiers left globally. Because robots do not have access to a shared map that tracks global frontiers, robot $i$ does not have a direct estimate of how much of the environment has been collectively explored by the team at any given timestep. 

\noindent \textbf{Challenge 2} \emph{Each robot $i$ in the team needs to asynchronously determine when to terminate exploration, using only its local information, by estimating total coverage of the environment.} 

This essentially prevents a robot from exploring the entire environment before terminating exploration and ensures that robots successfully map only a portion of the environment during normal operation. However, beyond normal operation, a situation may arise that would lead to heterogeneity in robot behavior. We specifically look at two heterogeneous behaviors in the team: i) varying speed of robots during exploration that can result from navigation challenges, and ii) complete failure of a robot leading to loss of it map data. We summarize this challenge as follows:

\noindent \textbf{Challenge 3} \emph{Under heterogeneous robot behaviors or failures, the robot team should dynamically adapt to ensure that the environment is fully explored.}

\floatstyle{spaceruled}
\restylefloat{algorithm}
\begin{algorithm}[t]
\caption{Frontier Exploration Algorithm \label{alg:FE}}
\begin{algorithmic}
\Require 
\\ Local frontier set $\textbf{F}_i$ of robot $i$
\\ Bearing measurements $\Phi$ (Angle-of-Arrival profile)
\\ Distance measurements $D$ (from range sensor)
\\ Current position estimate $\hat{x}_{ij}(t)$ of closest robot $j \in \mathcal{N}_i$ 
    \smallskip
    \While{Selecting Next Frontier}
    \State $\mathbf{\hat{\mathcal{Z}}_{ij}(t)} \gets [{\text{AverageDist}(D),\Phi}]$
    \State $\hat{x}_{ij} \gets \text{PDAF} (\bold{\hat{\mathcal{Z}_j}(t)})$ 
    \Comment{See Section \ref{sec:SAR_UWB}} 
    \State \textit{InsertInHgrid}($\hat{x}_{ij}$)
    \State $\textit{\textbf{occ-per}} \gets \text{HGrid occupancy percentage}$   
    \smallskip
    \For {$k= 0$ to $|{\textbf{F}}_i|$}
    \State $U_{\textbf{F}_i[k]} = 0$
        \For {each viewpoint $x_{v}^{\textbf{F}_i[k]}$}
            \State $\hat{X}_v^{\textbf{F}_i[k]} \gets $ \textit{RetrieveFromHGrid}($x_v^{\textbf{F}_i[k]}$)
            \State $\mathcal{I}_v^{\textbf{F}_i[k]} \gets$ \textit{InformationGain}
            \Comment{\text{Eqn: 
            \ref{eqn:information_gain}}}
            \State $\mathcal{C}_{\textbf{F}_i[k]} \gets \text{Path length from local planner}$
            \State $U_{i}^{\textbf{F}_i[k]} = \max( I_{v}^{\textbf{F}_i[k]}\textfractionsolidus{\mathcal{C}_{\textbf{F}_i[k]}}, U_{i}^{\textbf{F}_i[k]})$
        \EndFor
        \If{\textit{occ-per} \hspace{-0.1in} $>$ \hspace{-0.1in} \textit{\textbf{Soft\_Thresh}} \hspace{-0.08in} \& \hspace{-0.08in} $L(.)/S(.))\hspace{-0.02in}>\hspace{-0.02in}0.9$} 
            \State Remove $\textbf{F}_i[k]$ from $\textbf{F}_i$
            \Else 
        \State $\textbf{U}_{i} = [\textbf{U}_{i}, U_{i}^{\textbf{F}_i[k]}]$
        \EndIf
    \EndFor
    \smallskip
    \State Terminate exploration \textbf{if} $|\textbf{F}_i|$ == 0 \textbf{else}
    \State $\text{Visit next frontier}$ $\textbf{F}_i[k^*] \hspace{0.05in}\text{such that}\hspace{0.05in} U_{i}^{\textbf{F}_i[k]} = \max(\textbf{U}_{i})$
\EndWhile
\end{algorithmic}
\end{algorithm}
\section{Approach}\label{sec:approach}
\vspace{-0.05in}
This section outlines our approach to address the problem outlined in Section ~\ref{sec:problem}. In Section~\ref{sec:FE}, we assume that any robot $i\in\mathcal{R}$ can locally estimate relative position estimates $\hat{x}_{ij}$ for $j \in \mathcal{N}_i$. We then explain how each robot updates frontier information gain given these relative position estimates, leverages \textit{HGrid}, a quadtree-like data structure for efficient storage and retrieval, and achieve asynchronous exploration termination. Section~\ref{sec:heterogeneity_behavior_appraoch} explains how the team achieves full exploration while accounting for behavioral heterogeneity. Section~\ref{sec:SAR_UWB} describes how the relative position estimates $\hat{x}_{ij}$ are obtained on real robots using just wireless signal pings. 

\begin{figure}[h]
    \vspace{-0.125in}
    \centering       
    \includegraphics[scale=0.925]{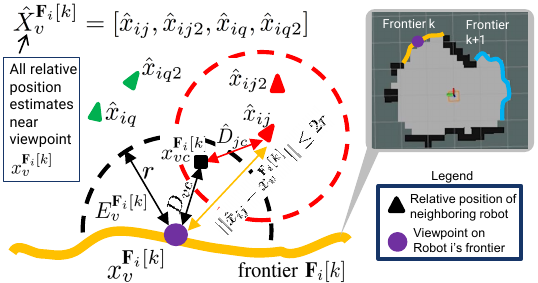} 
    \vspace{-0.1in}
    \caption{Schematic showing information gain computation}
    \label{fig:info_gain_overlap}
    \vspace{-0.32in}
\end{figure}

\begin{figure*}[t!] 
    \centering 
    \vspace{0.075in}
    \includegraphics[scale=0.5]{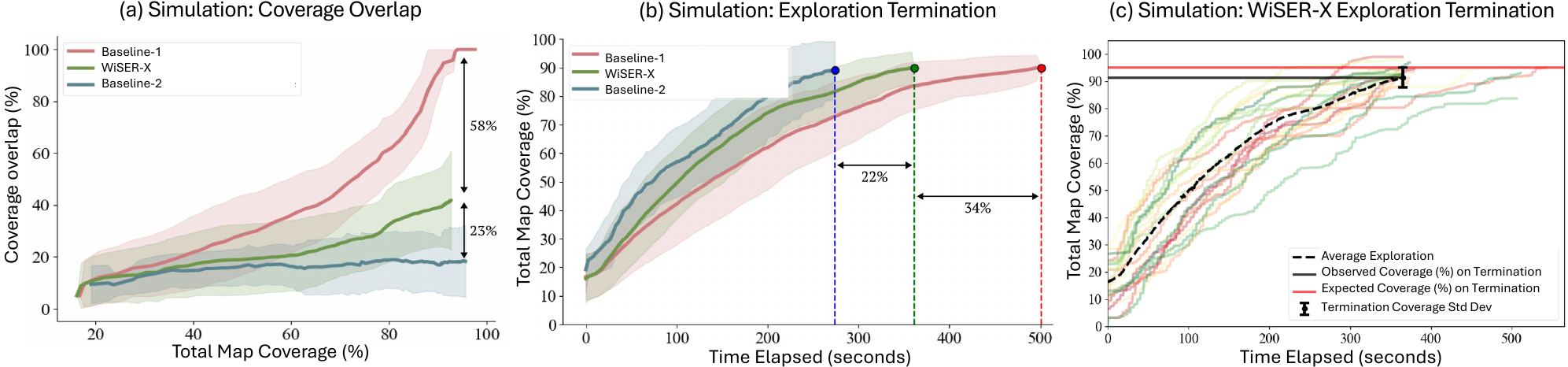}      
    \vspace{-0.05in}
    \caption{Simulation results that demonstrate performance of our algorithm against two baselines (no-information sharing baseline-1 and all-information sharing baseline-2). For the baseline algorithms, termination occurs when the merged map reached 95\% coverage. WiSER-X automatically triggers exploration termination when no valid frontiers are left. Plot (c) shows results for WiSER-X over 20 trials of simulation. Total map coverage is obtained from the map-merging Oracle and is used only for evaluation in case of WiSER-X algorithm.}
    \label{fig:sim_aggregate}
    \vspace{-0.25in}
\end{figure*}

\vspace{-0.05in}
\subsection{ WiSER-X Algorithm For Exploration}
\label{sec:FE}
\vspace{-0.05in}
WiSER-X uses a frontier-based approach (Algorithm~\ref{alg:FE}) as a local exploration algorithm for each robot. We follow the approach in \cite{Zhou2020FUELFU} when processing a frontier. Large frontiers, greater than LiDAR scan range $r$ in our case, are split into smaller segments along their principle axis obtained using Principal Component Analysis. To improve frontier selection, the utility is computed at three distinct grid cells i.e. \textit{viewpoints} on the frontier, corresponding to its center and the two extremes. $x_{v}^{\textbf{F}_i[k]}$ denotes the position of the viewpoint $v$ on $\textbf{F}_i[k]$. At each timestep $t$, robot $i$ calculates the utility $U_{\textbf{F}_i[k]}$ of all frontiers in its local map $\mathcal{M}_i$ while accounting for overlaps based on the estimated positions $\hat{x}_{ij}$.  

\subsubsection{Information Gain Computation}
For a robot $i$'s map, let $x_{v}^{\textbf{F}_i[k]}$ denote the position of viewpoint $v$ on frontier $\textbf{F}_i[k]$. $E^{\textbf{F}_i[k]}_v$ denotes the set of unexplored grid cells c within distance $D_{vc}\leq r$ around $v$  such that $D_{vc} = ||x_{v}^{\textbf{F}_i[k]} - x^{\textbf{F}_i[k]}_{vc}||^2$ where  
$x^{\textbf{F}_i[k]}_{vc}$ corresponds to the position of the grid cell c. $E^{\textbf{F}_i[k]}_v$ basically denotes the new area that robot $i$ would explore if it chooses to visit the frontier $\textbf{F}_i[k]$. We follow the development in \cite{PelzterFIGOP} and compute a grid cell's information using the sigmoid function $S(.)$ that reduces the information gain for cells farther away, essentially capturing the uncertainty in LiDAR range sensor measurements. 
\vspace{-0.01in}
\begin{align}
    & S(x_{v}^{\textbf{F}_i[k]}, x^{\textbf{F}_i[k]}_{vc})=\frac{1}{1+e^{(D_{vc} - \kappa_1)/\kappa_2}}
    \vspace{-0.2in}
\end{align}
where, $\kappa_1$ controls the midpoint and $\kappa_2$ controls the steepness of the curve, favoring closer cells but permitting distant cells to contribute as well. When deciding which frontier to visit next, robot $i$ queries its local HGrid to find relative positions $\hat{x}_{ij}$. These positions correspond to those of another robot $j$ in robot $i$'s neighborhood $\mathcal{N}_i$ that would indicate potential overlap with the unexplored area around viewpoint $v$ of robot $i$'s frontier $\textbf{F}_i[k]$ (See Fig. \ref{fig:info_gain_overlap}). Thus, $\hat{X}_v^{\textbf{F}_i[k]}$ denotes this set of $\hat{x}_{ij}$ around $v$ such that $||\hat{x}_{ij} - x_v^{\textbf{F}_i[k]}|| \leq 2r$. We estimate the information loss $ L(.)$ for a grid cell as follows:
\begin{equation}
    \begin{split}
    & L(\hat{X}_v^{\textbf{F}_i[k]},x^{\textbf{F}_i[k]}_{vc}) = \\ &\sum_{\hat{x}_{ij} \in \hat{X}_v^{\textbf{F}_i[k]}} \hspace{-0.125in} min(1, 1/\text{Tr}[\Sigma(\hat{x}_{ij})]) \cdot S(\hat{x}_{ij},x^{\textbf{F}_i[k]}_{vc})
    \label{eqn:information_loss}
    \vspace{-0.25in}
    \end{split}
\end{equation}
\noindent Here $S(.)$ uses $\hat{D}_{jc} = ||\hat{x}_{ij} - x^{\textbf{F}_i[k]}_{vc}||$ i.e., computing the information gain at grid cell c for neighboring robot $j$; $\min(1, 1/\text{Tr}[\Sigma(\hat{x}_{ij})])$ scales it using the trace of covariance $\Sigma$ for the position estimate $\hat{x}_{ij}$ accrued over robot $i$'s estimated trajectory. Intuitively, $\text{Tr}[\Sigma(\hat{x}_{ij})])$ scales down the information loss when the uncertainty of $\hat{x}_{ij}$ is higher. Equation~(\ref{eqn:information_loss}) thus computes the overlap around $\textbf{F}_i[k]$ based on the accuracy of the relative position estimate. To factor in the latest position of the closest robot $j$ in the choice of robot $i$'s frontier selection, we define $\beta$ as follows:
\begin{align}
    \vspace{-0.5in}
    \beta = \log_{10}(\min(|| \hat{x}_{ij}(t) - x_{v}^{\textbf{F}_i[k]}||_2))
    \label{eqn:beta_parameter}
    \vspace{-0.15in}
\end{align} 
where $\hat{x}_{ij}(t)$ denotes the relative position of any neighboring robot $j$ that is closest to the viewpoint position $x_{v}^{\textbf{F}_i[k]}$ at current timestep $t$. The total information gain $\mathcal{I}_{v}^{\textbf{F}_i[k]}$ at the viewpoint $v$ is given by: 
\begin{equation}
    \begin{split}
    &\mathcal{I}_{v}^{\textbf{F}_i[k]}\hspace{-0.05in} = \hspace{-0.05in}
    \beta \hspace{-0.12in} \sum_{c \in E^{\textbf{F}_i[k]}_v} \hspace{-0.12in} \max(0,\hspace{-0.02in} S(x_{v}^{\textbf{F}_i[k]}, x^{\textbf{F}_i[k]}_{vc}) \hspace{-0.03in}-  \hspace{-0.03in}L(\hat{X}_v^{\textbf{F}_i[k]},\hspace{-0.02in} x^{\textbf{F}_i[k]}_{vc}))
    \label{eqn:information_gain}
    \end{split}
\end{equation}
\vspace{-0.15in}

\begin{figure*}[t!] 
    \centering 
    \vspace{0.075in}
    \includegraphics[scale=0.525]{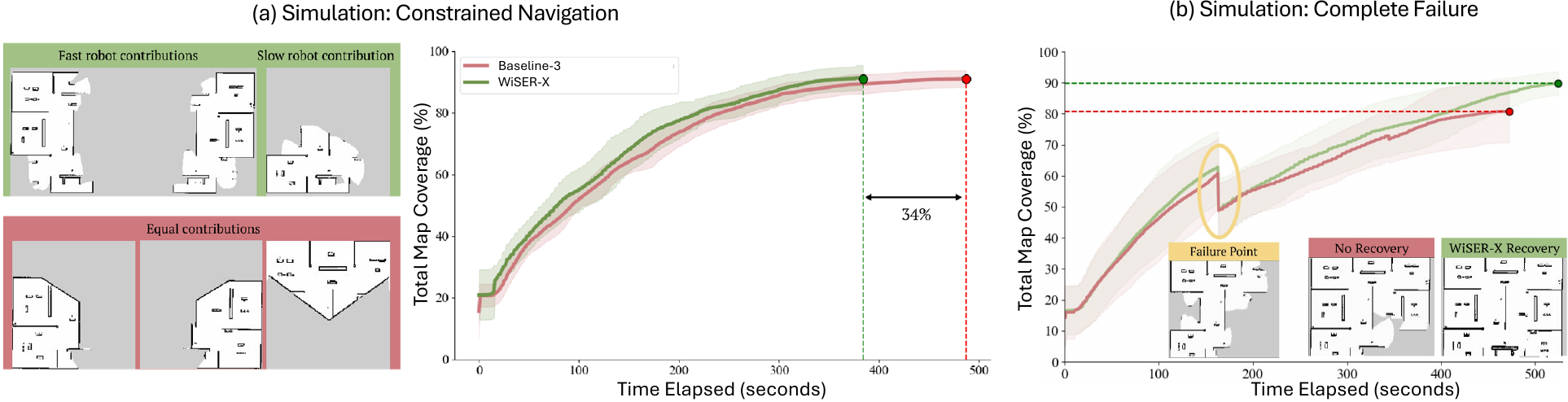}      
    \vspace{-0.05in}
    \caption{Simulation results for heterogeneous performance scenarios over 20 trials for each scenario. a) Shows map coverage over time for WiSER-X and Divide-and-Conquer Baseline-3 for one slow moving robot to emulate heterogeneous behavior resulting from challenging navigation. WiSER-X reduces average termination time by 140 seconds (34\%) while maintaining the same total coverage of the environment. b) Aggregate results and instance of simulation showing map coverage at termination time for WiSER-X after after a randomly chosen robot fails (loss of all map data from that robot, indicated in red in the images). After incorporating recovery behavior, WiSER-X enables other robots in the team to remap the area.}    \label{fig:sim_heterogeneity_aggregate}
    \vspace{-0.25in}
\end{figure*}

\noindent The max function ensures that the information gain per grid cell is non-negative and lower bounded to 0. $\beta$ scales the net information gain based on the proximity of the nearest neighboring robot $j$ to viewpoint $v$, thereby incentivizing the robots to spread out. $\mathcal{C}_{\textbf{F}_i[k]}$ denotes the navigation cost corresponding to the actual path length between the robot's current position $x_i$ and the frontier center, as computed by the local path planner. We note that the difference between path length to different viewpoints is trivial and hence not computed separately.
$\mathcal{I}_{v}^{\textbf{F}_i[k]}\textfractionsolidus{\mathcal{C}_{\textbf{F}_i[k]}}$ denotes the utility for a viewpoint and the maximum value among all viewpoint utilities is assigned to $U_{\textbf{F}_i[k]}$. Robot $i$ computes $U_{\textbf{F}_i[k]}$ for all frontiers in $\textbf{F}_i$ and navigates to the frontier with the highest utility. In order to avoid oscillatory behavior, where a robot might move back-and-forth between two frontiers of similar utility, WiSER-X ensures that the robot commits to its chosen frontier until it is at least halfway to said frontier, based on the path length. Only at this point does the robot reevaluate the utility of all the frontiers generated in the current timestep and change the target frontier if required.

\subsubsection{HGrid operations}
\label{sec:hgrid}
Robot $i$ stores its own position $x_i$, the estimated position of neighboring robots $\hat{x}_{ij}$, and the trace of the covariance matrix $\text{Tr}[\Sigma(\hat{x}_{ij})]$, which represents the uncertainty of the estimated positions. These observations are collected over the duration of exploration and stored in the HGrid, a discretized representation of the environment local to each robot and has a structure similar to a quadtree to enable information retrieval, such as $\hat{X}_v^{\textbf{F}_i[k]}$, in $O(\log N)$ time~\cite{Finkel1974QuadTA}. Each HGrid cell has dimensions equivalent to twice the sensor range $r$ (see Fig~\ref{fig:intro_fig}).

A robot's HGrid cell represents a local record of coverage and maintains the count of how many times it had been visited by itself or other robots. An HGrid-cell is marked as \textit{filled} once the count exceeds a user-defined threshold, indicating that the area had been sufficiently covered. 

\subsubsection{ Exploration termination}
\label{sec:termination}
As the robot team's collective coverage of the environment increases, the information gain of a robot $i$'s frontiers gradually decreases. Using its HGrid, robot $i$ continuously updates its coverage estimation of the environment. We define invalid frontiers as those for which the percentage of information loss (Eqn.~\ref{eqn:information_loss}) over information gain (without considering any overlaps, obtained by setting $L(.) =0$ in Eqn.~\ref{eqn:information_gain}) is greater than 90\%. WiSER-X triggers the termination behavior by setting a \textit{soft} threshold limit on the HGrid's occupancy. When the occupancy reaches the soft threshold, robot $i$ disregards invalid frontiers and terminates exploration asynchronously if no valid frontiers are generated in the current timestep. This behavior ensures that robots do not stop prematurely.

\subsection{Adapting to heterogeneous behaviors of robots}
\label{sec:heterogeneity_behavior_appraoch}
\vspace{-0.05in}
We consider two heterogeneous behaviors for the robots. The first scenario involves varying the exploration speed where some robots explore the environment faster than others. It simulates a scenario where some robots may end up in areas of the environment that make navigation more challenging. WiSER-X automatically addresses such behavior by directing the faster robots to cover more area, compensating for the reduced performance of the slow robots.

The second scenario simulates complete failure of a robot, resulting in the loss of access to its local map for the areas it has covered. This necessitates that the rest of the robots in the team re-explore the areas covered by the failed robot to ensure complete exploration of the environment. $\tau_{ij} \in [0,1]$ indicates whether robot $i$ detects that any robot $j \in \mathcal{N}_i$ has failed $(0)$ or is functional $(1)$. We assume that when a robot $j$ fails, robot $i$ can no longer obtain ping packets from it and thus asynchronously sets $\tau_{ij} = 0$. Given the relative position $\hat{x}_{ij}$ of robot $j$, the function $\textbf{T}(.)$ maps it to $\tau_{ij}$. As $\tau_{ij}$ is maintained as a pointer this operation is executed in $O(1)$. We thus enable the re-exploration behavior by updating equation~(\ref{eqn:information_loss}) as follows:
\begin{equation}
    \begin{split}
    & L(\hat{X}_v^{\textbf{F}_i[k]},x^{\textbf{F}_i[k]}_{vc}) = \\ &\sum_{\hat{x}_{ij} \in \hat{X}_v^{\textbf{F}_i[k]}} \hspace{-0.125in}  \textbf{T}(\hat{x}_{ij}) \cdot min(1, 1/\text{Tr}[\Sigma(\hat{x}_{ij})]) \cdot S(\hat{x}_{ij},x^{\textbf{F}_i[k]}_{vc})
    \label{eqn:information_loss_with_tau}
    \vspace{-0.25in}
    \end{split}
\end{equation}
Thus, when the robot $i$ computes the set of frontiers $\textbf{F}_i$ in the next timestep, it disregards any information loss due to the failed robot $j$ and enables ``re-exploration" of the areas robot $j$ visited prior.

\subsection{Relative Positions using onboard sensing}\label{sec:SAR_UWB}
\vspace{-0.05in}
Previous sections assume that the robots can locally obtain relative positions $\hat{x}_{ij}$. In this section we discuss how to obtain we demonstrate our algorithm on real hardware using RF signals using $\hat{x}_{ij}$ from noisy measurements. Specifically, we use the physical properties of signals emitted by commodity hardware such as WiFi cards and Ultra-wide band modules (UWB) to estimate $\hat{x}_{ij}$. Robot $i$ fuses data from its single onboard UWB sensor (estimating the range from signal time-of-flight) and the WSR toolbox~\cite{WSRToolbox} that uses WiFi (using signal phase to estimate bearing).
\subsubsection{Range estimation} We use the opensource code~\cite{zephyr-dwm1001} to obtain range measurements from the UWB module.
\subsubsection{Bearing estimation} The WSR Toolbox uses the relative signal phase data between two receiving antennas connected to a WiFi card to eliminate phase noise due to carrier frequency offset~\cite{Kumar2014AccurateIL}. It emulates a ``virtual antenna array" which captures all signal paths transmitted between robots. To enable emulation of this virtual antenna array, the two-antenna setup is deployed on a servo that rotates back-and-forth, obtaining an Angle-of-Arrival (AOA) profile measurement every three seconds onboard the robot ~(Fig. \ref{fig:intro_fig}B) . The AOA profile captures all signal paths; only one corresponds to the true bearing while other measurements correspond to signal multipaths, the reflected and attenuated paths as the signals propagate through the environment~\cite{Xiong2013ArrayTrackAF}. The WSR toolbox returns \textit{top N} peak angles which are at least 10\% of the strongest peak in the AOA profile

\subsubsection{Estimation of relative position} The \textit{top N} peak returned by the WSR Toolbox, along with the average range measurement from UWB are combined to form a set of measurement inputs $\bold{\hat{\mathcal{Z}}_{ij}(t)}$. We use the PDAF~\cite{ShalomPDAF} for continuous estimation of $\hat{x}_{ij}$ using $\bold{\hat{\mathcal{Z}}_{ij}(t)}$ and briefly summarize its implementation. PDAF follows the Kalman Filter (KF) predict and update loop while also accounting for ambiguity in measurements, for example, signal multipath angles in our bearing measurements. For each range and bearing measurement pair in $\bold{\hat{\mathcal{Z}}_{ij}(t)}$, PDAF computes the Mahalanobis distance of the residual error and uses a gating threshold value of 95\% to reject measurements that are more likely to be from multipaths. It assigns a likelihood to the rest of the measurements based on how well they match the predicted location of the target. The filter then uses a weighted combination of all valid measurements and uses the KF update to estimate the position of robot $j$. Our opensource code provides complete implementation for reproducibility.

Thus, by leveraging inter-robot onboard relative position estimates using the physical properties of wireless signals, our algorithm allows for coordination between robots without requiring explicit information exchange.
\begin{figure}[t!] 
    \centering 
    \vspace{0.075in}
    \includegraphics[scale=0.5225]{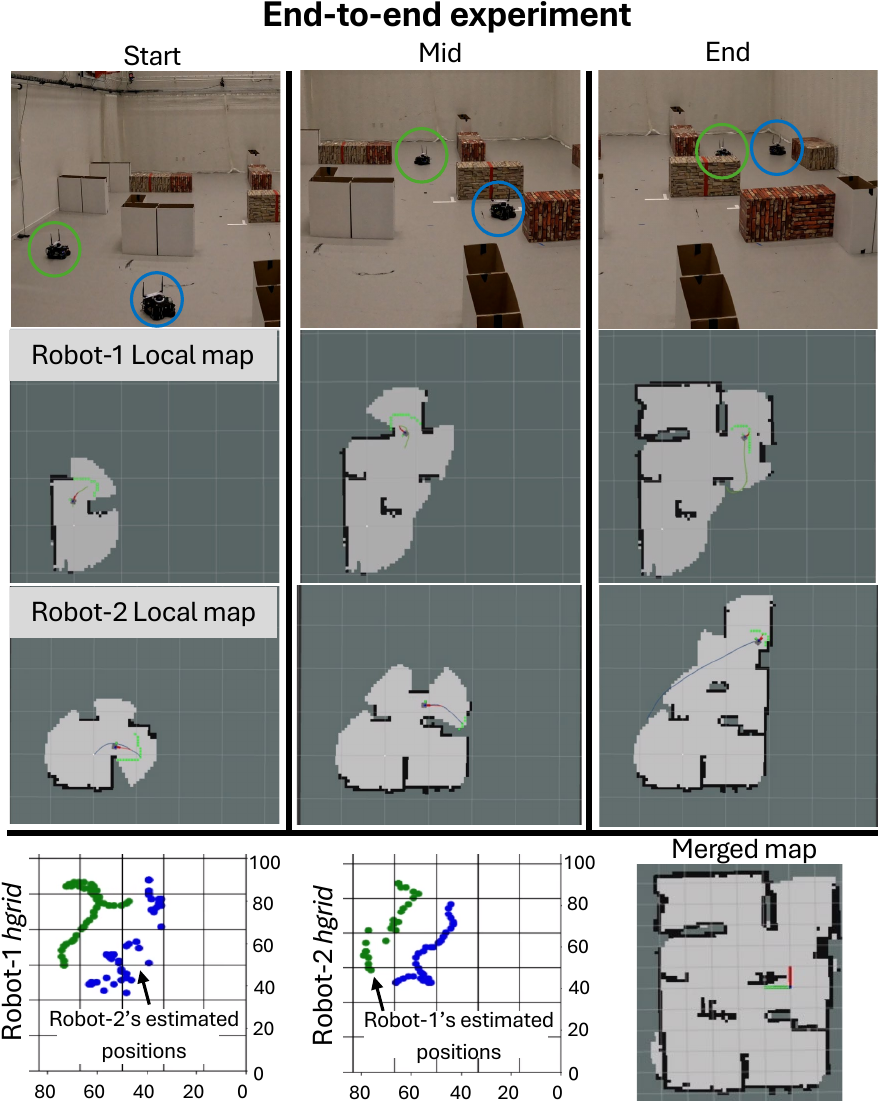}  
    \vspace{-0.05in}
    \caption{Qualitative results for a trial of the end-to-end hardware experiment showing the state of the exploration at three instances.}
    \label{fig:hw_end_to_end}
    \vspace{-0.25in}
\end{figure}

\section{Results}\label{sec:results}
\vspace{-0.05in}
We first validate WiSER-X through extensive simulation experiments and compare its performance against three baseline algorithms. Next, we conduct hardware experiments that demonstrate decentralized exploration using two mobile ground robots with all onboard sensing.

\subsection{Evaluation metrics}
We evaluate the performance of our algorithm and comparison baselines using a global \textit{map-merging oracle} ROS package~\cite{Horner2016} and the following metrics:
\begin{itemize}[leftmargin=0.125in]
    \item \textbf{Coverage overlap}: The overlap between individual robot maps is measured to evaluate the algorithm's effectiveness in reducing redundancy.
    \item \textbf{Termination time}: The evaluation focuses on how quickly and consistently the algorithm enables asynchronous termination of exploration.
    \item \textbf{Heterogeneity in robot behavior}: We test the algorithm's robustness in scenarios pertaining to variability in the robot team's performance by simulating complete failure of a robot and reducing its speed to emulate constrained navigation. 
\end{itemize}

\subsection{Simulation Experiments}
\subsubsection{Environment}
Our Gazebo simulations used a 1600 $m^2$ cluttered office environment with three ground robots, each equipped with an on-board LiDAR. Each robot was initialized at random locations throughout the environment. WiSER-X is implemented on top of the C++ explore-lite~\cite{Horner2016} package. For the WiSER-X algorithm, we generated noisy range and AOA estimates using the true positions of the robots with added noise. We added a zero mean Gaussian noise and stddev of 10 cm for range measurements and 5 degrees for bearing measurements~\cite{Wang2019ActiveRF}. Our user defined threshold to mark an HGrid-cell as filled is set to three to account for any spurious observations of relative positions. We conduct 20 trials for each evaluation scenario.

\subsubsection{Comparison baselines}
\begin{itemize}[leftmargin=0.125in]
    \item \textbf{Baseline-1: Independent Exploration}: A frontier-based exploration algorithm based on the explore-lite package~\cite{Horner2016}. In this baseline, each robot selects frontiers, without considering the relative position of other robots. This represents a zero-information-sharing exploration strategy.

    \item \textbf{Baseline-2: Full Information Exchange}: A global frontier-based algorithm using the Rapidly-Exploring Random Tree (RRT)-exploration package \cite{Mukhopadhyay2024}. This package employs RRT~\cite{LaValle1998RapidlyexploringRT}, incrementally building a tree from a starting point by randomly sampling points in the space and expanding the tree towards those points, favoring high-utility goal points. A global “assigner” node allocates frontiers to robots as an oracle system, representing a full-information-sharing exploration strategy.
    \item \textbf{Baseline-3: Divide-and-Conquer}: Each robot is assigned a specific area to map and end exploration once that area is completely explored. This baseline is only used to evaluate heterogeneous robot behaviors.
\end{itemize}
\noindent For Baseline-1 and 2, the map-merging oracle is used to terminate exploration, enacted at approximately 95\% completion of overall exploration. WiSER-X performed termination independently for each robot without this central server involvement, as described in \ref{sec:termination}.

\begin{figure}[t!]
    \centering 
    \vspace{0.075in}
    \includegraphics[scale=0.5225]{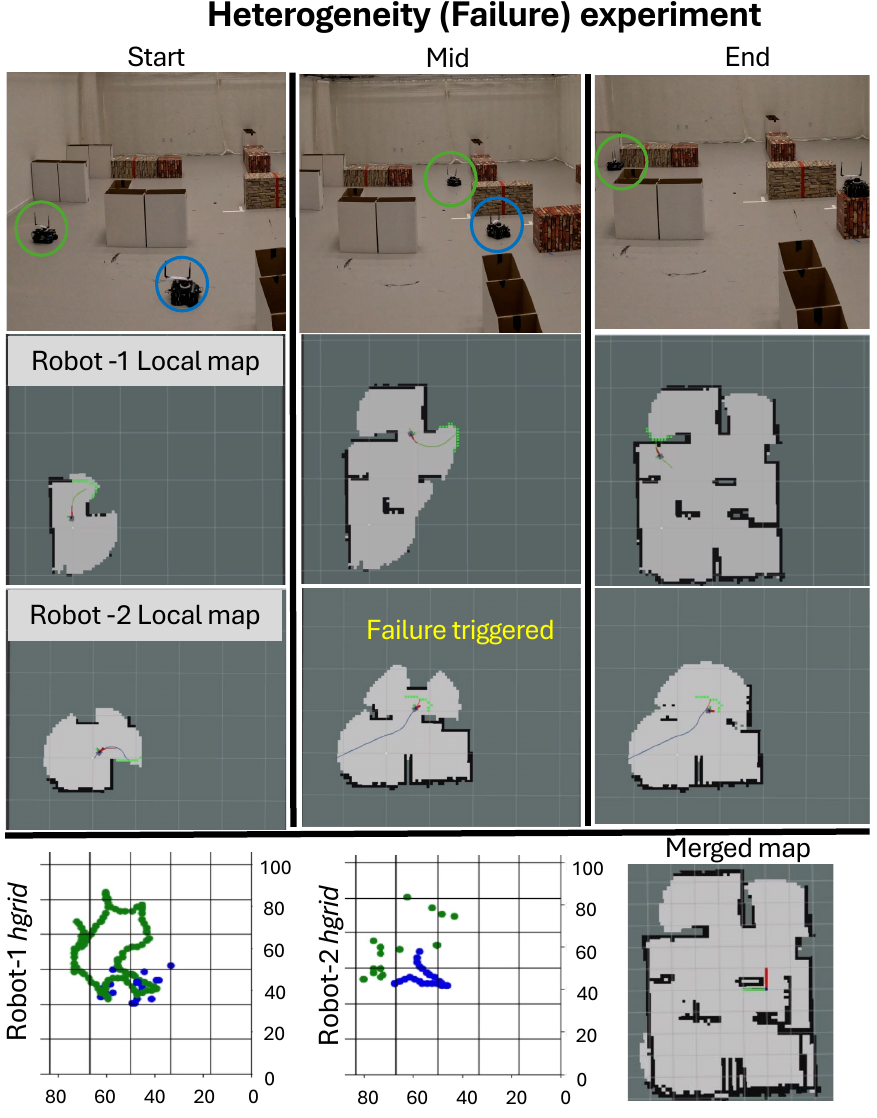}
    \vspace{-0.05in}
    \caption{Qualitative results for a trial of complete failure scenario demonstrating the remapping behavior.}
    \label{fig:hw_failure_qual}
     \vspace{-0.25in}
\end{figure}

\subsubsection{Coverage overlap and termination time} 
\noindent \textbf{Coverage overlap.} As shown in Figure \ref{fig:sim_aggregate} (a), WiSER-X reduces the mean coverage overlap at the end of exploration by 58\% compared to the zero-information-sharing Baseline-1 and only 23\% more overlap than the full-information-sharing Baseline-2. \textbf{Termination time.} Figure \ref{fig:sim_aggregate} (b) shows that WiSER-X terminates, on average over 20 trials, 1.65X faster than Baseline-1 in simulation, saving approximately 34\% time. Although both algorithms terminate after Baseline-2, WiSER-X takes 22\% longer while Baseline-2 takes 56\% longer. Figure \ref{fig:sim_aggregate} (c) illustrates WiSER-X's exploration termination time. The graph shows that, the algorithm enables robots to asynchronously terminate exploration, on average, at 93\% total map coverage (3.6\% standard deviation) in approximately 365 seconds (77.5 second standard deviation). 

\begin{figure*}[t!] 
    \centering 
    \vspace{0.075in}
    \includegraphics[scale=0.5]{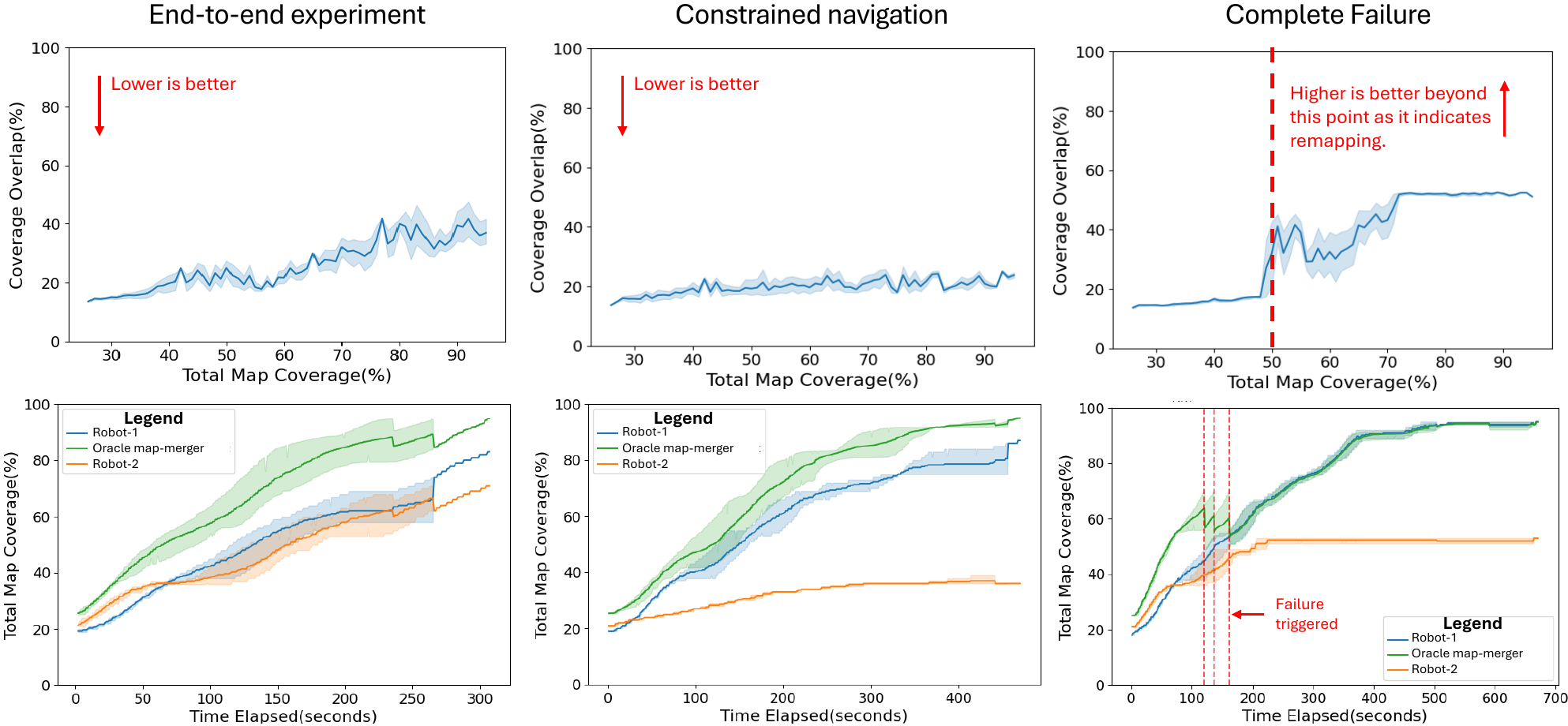}      
    \vspace{-0.05in}
    \caption{Results for hardware experiments with all onboard sensing, aggregated over three trials for each scenario. For the end-to-end experiment both robots continuously map the environment. During constrained navigation, robot-2 moves slower than robot-1, while complete failure simulates a failure scenario leading to all loss of data for robot-2.}
    \label{fig:hw_all_aggregate}
    \vspace{-0.275in}
\end{figure*}

\subsubsection{Heterogeneity in robot behavior}
To simulate constrained navigation, one of the three robots was assigned a slower speed, reducing its coverage and increasing total exploration time (Figure~\ref{fig:sim_heterogeneity_aggregate}a). WiSER-X compensates automatically by allowing faster robots to cover more ground, reducing average termination time across 20 trials by 34\% (mean of 140 seconds with a std dev of 9 seconds). 

A second scenario simulated complete robot failure. One robot was randomly disabled after 50–70\% of total exploration (based on the map-merger oracle). This timing allowed for a clear observation of the WiSER-X algorithm with and without remapping behavior. In this failure scenario, WiSER-X successfully recovers approximately 10\%, (with a 4.25\% std dev), of the map that otherwise would have been lost due to a robot's failure. WiSER-X's implicit coordination allows for dynamic adaptation to robot failures and reallocation of exploration tasks in real time. Figure~\ref{fig:sim_heterogeneity_aggregate}(b) shows the aggregate results for this scenario.

\subsection{Hardware Experiments}
We conducted hardware experiments in a 70 m$^2$
indoor environment (Figure~\ref{fig:intro_fig}) using two customized TurtleBots equipped with UP-Squared compute boards. Each robot uses the GMapping SLAM algorithm~\cite{Grisetti2007Gmapping} and the TEB planner~\cite{Rsmann2017IntegratedOTTEBPlanner} for local navigation. Range and AOA profile measurements were obtained from Qorvo DWM1001-DEV UWB modules and a WiFi-based WSR toolbox, respectively. The UWB module streamed range estimates at 10 Hz, while the WSR toolbox processed approximately 100 ping packets in total collected over a three second sampling window, simultaneously yielding an updated AOA profile every two seconds via a sliding-window approach. All sensing, measurement generation, and SLAM computations were performed onboard, whereas the WiSER-X algorithm was executed offboard in realtime to simplify evaluation. The map-merging oracle was used only for evaluation, and robots did not use the merged-map to coordinate their exploration.

These experiments evaluate WiSER-X for each of the following scenarios - End-to-End experiment with all onboard sensing, heterogeneity in navigation, and complete failure of a robot. Figure ~\ref{fig:hw_end_to_end} shows qualitative results for a single trial of the end-to-end scenario where each robot has approximately covered half of the environment. An example of the complete failure scenario is shown in Figure~\ref{fig:hw_failure_qual} where robot-1 remaps the area explored by robot-2 when a complete failure is detected (no ping packets received from robot-2). Figure~\ref{fig:hw_all_aggregate} shows the aggregate results across three trials for each of these scenarios. The first row compares overlap between the local maps of the robots and the second row shows the exploration duration. WiSER-X minimizes overlap during the end-to-end experiment and is able to adapt the exploration strategy of individual robots to address heterogeneous performance.

\section{Conclusion}\label{sec:conclusion}
\vspace{-0.05in}
This paper demonstrates that wireless signal sensing can induce global coordination behaviors through entirely local, onboard algorithms without explicit information exchange. By extracting relative position information directly from signal ping packets, robots can coordinate effectively without sharing maps or other high-bandwidth data. This capability enables efficient multi-robot exploration in bandwidth-constrained environments, where each robot relies solely on local sensing over ping packets for coordination. Future work will extend WiSER-X to scenarios with intermittent communication, underwater acoustic sensing, and exploration tasks that incorporate imperfect prior maps, such as floor plans.


\bibliographystyle{IEEEtran}
\bibliography{root}

\end{document}